\pdfoutput=1

\documentclass[conference]{IEEEtran}

\usepackage[pdftex]{graphicx}

\begin{document}
%
\title{Mixed context networks for semantic segmentation}

\author{\IEEEauthorblockN{Haiming Sun}
\IEEEauthorblockA{Hikvision Research Institute\\
Email: sunhaiming@hikvision.com}
\and
\IEEEauthorblockN{Di Xie}
\IEEEauthorblockA{Hikvision Research Institute\\
Email: xiedi@hikvision.com}
\and
\IEEEauthorblockN{Shiliang Pu}
\IEEEauthorblockA{Hikvision Research Institute\\
Email: pushiliang@hikvision.com}}

\maketitle

\begin{abstract}
Semantic segmentation is challenging as it requires both object-level information and pixel-level accuracy. Recently, FCN-based systems gained great improvement in this area. Unlike classification networks, combining features of different layers plays an important role in these dense prediction models, as these features contains information of different levels. A number of models have been proposed to show how to use these features. However, what is the best architecture to make use of features of different layers is still a question. In this paper, we propose a module, called mixed context network, and show that our presented system outperforms most existing semantic segmentation systems by making use of this module.
\end{abstract}


\IEEEpeerreviewmaketitle

\section{Introduction}
Semantic segmentation is a classic problem in computer vision. It is challenging as it requires both object-level information and pixel-level accuracy, especially when there is a large variation in sizes of objects.\\
Recently, methods based Fully Convolutional Networks (FCN) \cite{long2015fully} significantly improve the performance in many datasets. FCNs are usually transformed from pre-trained classification networks (e.g., VGG-16, Resnet), by making use of ${1\times1}$ convolutional layers to replace fully-connected layers. The simplest FCN (e.g., FCN-32s) generate coarse score map. To gain fine predictions, one strategy is to combine fine predictions, which generated from early layers, like FCN-8s \cite{long2015fully}. \cite{hariharan2015hypercolumns} shows that FCN-8s can be seemed as an efficient implementation of linear prediction using the hypercolumn features, which are extracted from multiple layers of the network.\\

However, further studies show that the linear predictor on hypercolumn features might not be a good choice. \cite{pinheiro2016learning} uses a stage-wise refinement module to combine features from multiple layers. \cite{bansal2016pixelnet} shows that a MLP defined on hypercolumn features lead to better results than linear prediction.\\

On the other hand, the dilated convolution, also called atrous convolution, plays an important role in dense prediction systems. The dilated convolution enlarges the receptive field without loss of resolution of prediction. FCN-based models are able to benefit from it, and generate fine score maps without combining features from multiple layers. Furthermore, DilatedNet \cite{chen2014semantic,yu2015multi} can arbitrarily enlarge the receptive field by adjusting the sampling rates of dilated convolutions. Such systems usually gain more accurate predictions as more context information is considered.\\

Dilated convolution enlarges the receptive field without loss of resolution of prediction. But keeping the resolution does not mean keeping detailed information. Combining features from multiple dilated convolutions with different sampling rates should be meaningful for dense prediction. To combine these features, one strategy is extending the refinement module \cite{pinheiro2016learning} to all dilated convolutional layers with different sampling rates, but it is cost if many dilated convolutional layers are added like models in \cite{yu2015multi}. Another strategy is use a module like atrous spatial pyramid pooling (ASPP) \cite{chen2016deeplab}, which fuses features from multiple parallel dilated convolutional layers with different sampling rates. But the sampling rates are fixed in ASPP, and you maybe need to do some experiments to choose good rates. The third strategy is use hypercolumn features and sparse predictions like in \cite{bansal2016pixelnet}, but it is not computational efficient when predicting.\\

In the paper, we propose a module called mixed context network, which mixes features of multiple dilated convolutions, and seems to have ability to learn to identify the most relevant scales. By using the mixed context network, our architecture produces competitive results in PASCAL VOC 2012 semantic segmentation challenge and MIT SceneParsing150 challenge.

\section{Related work}
Many systems have been proposed to handling scale variability in semantic segmentation. One of the most common approaches is extracting score maps from multiple rescaled versions of the original image by making use of parallel CNN branches \cite{chen2015attention}. \cite{long2015fully} uses skip connections to combine the predictions of fine layers and coarse layers. \cite{hariharan2015hypercolumns} gains improvements on detection and segmentation tasks by Hypercolumns representation which could be generated by skip connections. \cite{chen2016deeplab} attacks this problem by using atrous spatial pyramid pooling (ASPP), which is constructed by multiple parallel atrous convolutional layers with different sampling rates. \cite{drozdzal2016importance} studies the influence of both long and short skip connections, and finds that both of them are helpful to FCN-based method.

\section{Proposed method}
We first introduce the overall architecture at this section. After that, we will discuss several options for the mixed context network module, and a memory-efficient refinement module called message passing network will be proposed at last.

\begin{figure*}[!t]
\centering
\includegraphics[width=0.9\textwidth]{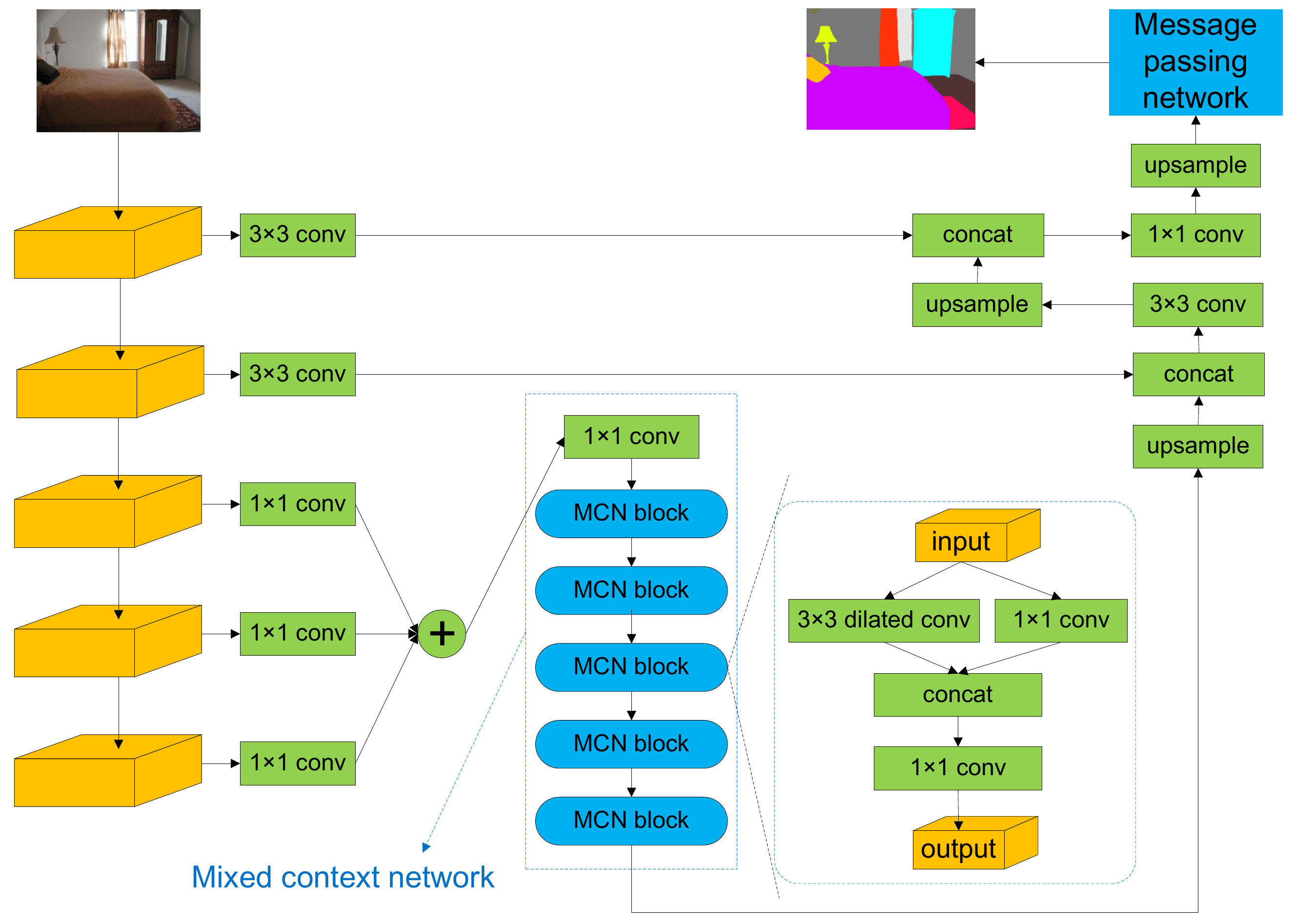}
\caption{Architecture for semantic segmentation. BN layers and non-linearity layers are omitted for brevity.}
\label{Fig.1}
\end{figure*}

\subsection{Overall Architecture}
Given a FCN-based model, such as the DilatedNet proposed by \cite{yu2015multi}, we consider effective ways to use the features from multiple layers. The DilatedNet can be divided into two parts. The bottom part is transformed from pre-trained classification networks, and the upper part is formed by a stack of dilated convolutions with different sampling rates. We deal with each part on its merits, as the two parts play different roles.\\

The structure of the bottom part usually stays the same with classification networks at least has two reasons. The first reason is that the segmentation models need to fine-tune from classification networks, as the pixel-level label annotations are expensive, and the data set is usually not big enough to train a deep model. The second reason is that the receptive fields of most classification networks are just the right size to segment most objects in normal size pictures.\\

Furthermore, a number of experiments have shown that enlarging the receptive fields of models leads to better results in most public datasets (e.g., PASCAL VOC). This is why the upper part is added, which usually formed by dilated convolutions. The receptive field could be arbitrarily enlarged by adjusting the sampling rates of dilated convolutions, but it is not easy to decide which rate is the best. On the other hand, it seems necessary to combine features from multiple dilated convolutional layers, as the dilated convolution leads to large variation in receptive field.\\

For the above reasons, we propose a model (Fig.\ref{Fig.1}) for dense prediction. We begin with casting classification convolutional networks to fully convolutional networks, and pick out several feature maps of FCN to be combined (one feature map per pooling layer). The feature maps from deeper layers, which encoding more high-level semantic information, are fused first, and fed into a module called mixed context network (MCN). The mixed context network mixes features of multiple dilated convolutions, and seems to have ability to learn to identify the most relevant scales. Then, refinement modules \cite{pinheiro2016learning} are stacked by making use of output feature map of MCN and feature maps from lower layers. At last, a module called message passing network is added to enforce spatial consistency across labels.\\

Compared with directly feeding the output score map of FCN into the context network \cite{yu2015multi}, our system gains significantly improvements by using combination of feature maps. For models like FCN-32s \cite{long2015fully}, feature maps should be up-sampled to the same resolution before fused, but for models which remove pooling layers by making use of dilated convolutions \cite{chen2014semantic}, the up-sampling operator can be omitted. To avoid too many parameters, reducing the number of channels of feature maps before fed into MCN is recommendatory for most tasks.\\

Which feature maps of the FCN to pick up also has an impact on system performance. Our experiments have shown that, feature maps from deeper layers lead to better result. So, taking VGG-16 \cite{simonyan2014very} as an example, feature maps from conv2\_2, conv3\_3, conv4\_3 and conv5\_3 layers work better than feature maps from pool1, pool2, pool3 and pool4 layers.

\begin{figure}[!t]
\centering
\includegraphics[width=0.5\textwidth]{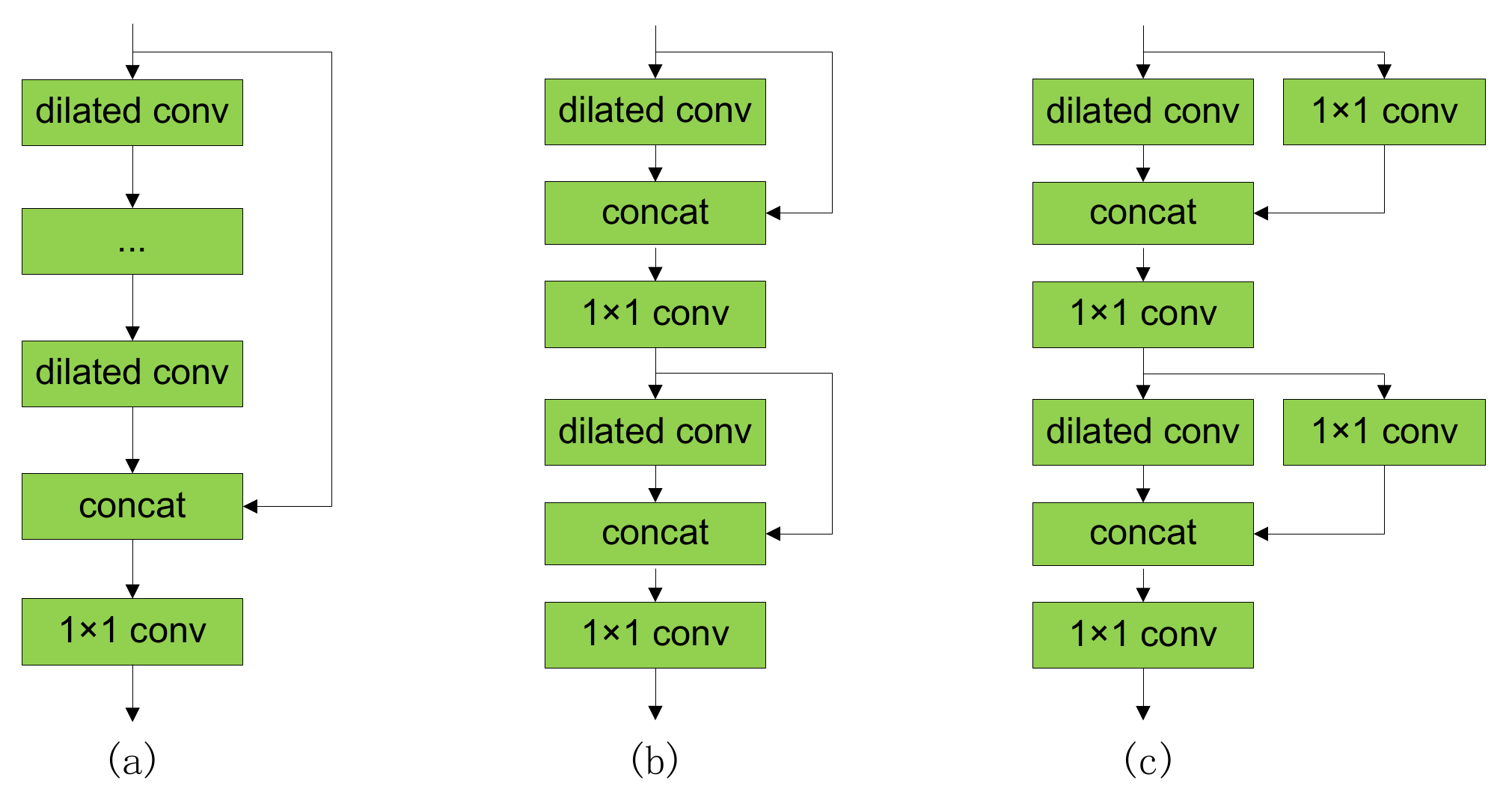}
\caption{(a) Dilated convolutions with a common long skip connection. (b) Dilated convolutions with common short skip connections. (c) Mixed context network composed of a stack of blocks, which contains two parallel convolutional layers: a dilated convolution layer and a ${1\times1}$ convolution layer. The feature maps from the two convolution layers are concatenated and fed into a ${1\times1}$ convolution layer to generate the output feature map.}
\label{Fig.2}
\end{figure}

\subsection{Mixed context networks}
In this paper, we consider three architectures for combining features of multiple dilated convolutional layers, which are shown in Fig.\ref{Fig.2}. Follow the work of \cite{yu2015multi}, kernel size of any dilated convolution in these models is set to 3, and we double the sampling rate layer by layer from bottom to top.\\

Skip connections are popular in dense prediction systems. Common skip connections just concatenate or fuse features from multiple layers of feedforward networks \cite{long2015fully,hariharan2015hypercolumns}. We design two architectures by making use of common skip connections. Given a stack of dilated convolutions, the first architecture (Fig.2 (a)) just concatenate the input feature maps of the first dilated convolution and the output feature maps of the last dilated convolution. In contrast, the second architecture (Fig.2 (b)) concatenate the input feature map and output feature map of one dilated convolution, and feed the concatenated features to the next dilated convolution. The intermediate ${1\times1}$ convolutional layer are used to adjust the number of channels of feature maps.\\

Furthermore, we design the mixed context network, in which, we do not directly use common skip connections, but add a convolutional layer before the features are concatenated. In the mixed context network, multi-scale features are mixed stepwise, and the network is able to learn to put different weights to features of different scales.

\begin{figure}[!t]
\centering
\includegraphics[width=0.5\textwidth]{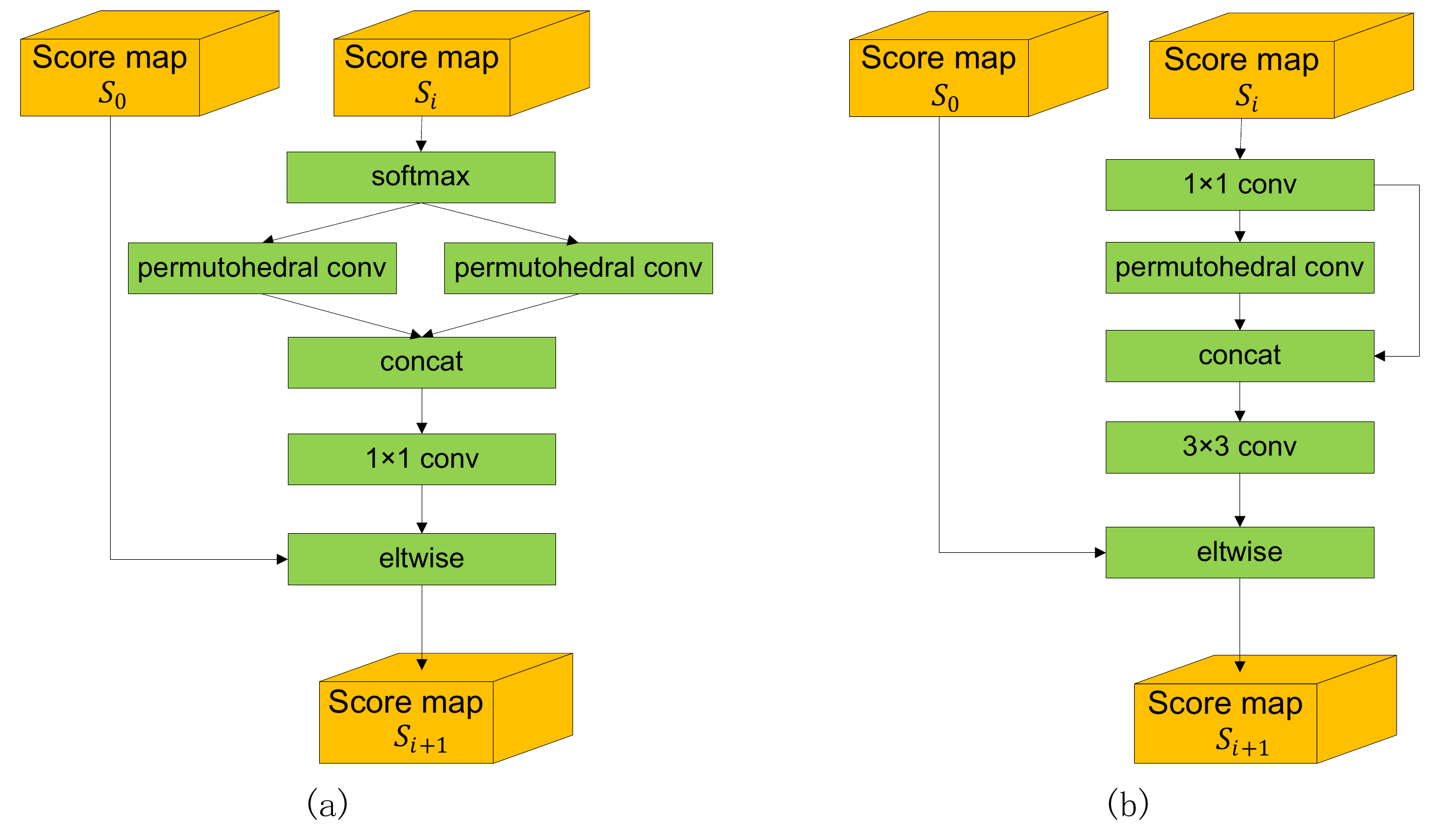}
\caption{(a) CRF-RNN, we merge the weighting filter outputs step and the compatibility transform step by making use of a ${1\times1}$ convolutional layer (b) MPN compresses the information of score map, and use a sub network to approximate simulate the message passing process. The ${1\times1}$ convolutional layer reduces the number of channels of score maps, and the ${3\times3}$ convolutional layer increases the number of channels of score maps.}
\label{Fig.3}
\end{figure}

\subsection{Message passing network}
As FCNs produce separate predictions for each pixel, most systems enforce spatial consistency across labels by making use of dense-CRF \cite{koltun2011efficient}. \cite{zheng2015conditional} shows that the mean-field CRF inference can be reformulated as a recurrent neural network, and the integrated system can be trained end-to-end with the back-propagation algorithm.\\

However, the CRF-RNN is memory cost, especially when the class number is large, for example 150 in ADE20K. One can decrease the resolution of input patches to save memory, but it is not benefit for training of deep layers, which have large receptive fields. We attack this problem by making use of a memory-efficient module called message passing network (MPN), shown in Fig.\ref{Fig.3}.\\

In each iteration of MPN, the number of channels of score map $S_{i}$ is first reduced from N to $N_{s}$, that $N_{s}$ is smaller than N. $S_{i}(i>0)$ is the output score map of last iteration, and $S_{0}$ is the input score map of MPN. The reduced score map is fed into a permutohedral convolutional layer, which performs high-dimension pairwise term. Compared to CRF-RNN, we drop the smooth pairwise term, and instead use a concatenation layer follows a ${3\times3}$ convolutional layer. The output score map $S_{i+1}$ is the sum of $S_{0}$ and the output feature maps of the ${3\times3}$ convolutional layer, which has exactly the same number of channels as score map $S_{0}$.

\section{Experiments}
Our implementation is based on the Caffe library \cite{jia2014caffe}. We trained all models using Nesterov. Mini-batch size is set to 20 and momentum is set to 0.9. Training was performed with an initial learning rate of 0.01, and multiply it by 0.1 every 50K iterations.

\begin{table}[!t]
\renewcommand{\arraystretch}{1.3}
\caption{Results of Validation Set of PASCAL VOC 2012}
\label{Table_1}
\centering
\begin{tabular}{c|c}
\hline
models & MeanIU\\
\hline
FCN\_256CT & 0.784\\
\hline
FCN\_256CT\_long\_skip & 0.793\\
\hline
FCN\_256CT\_short\_skip & 0.790\\
\hline
FCN\_MCN & 0.806\\
\hline
FCN\_MCN\_long\_skip & 0.802\\
\hline
\end{tabular}
\end{table}

\subsection{PASCAL VOC 2012}
We test our models on the PASCAL VOC 2012 semantic segmentation benchmark, consisting of 20 foreground object classes and one background class. All our models are trained on augment Pascal dataset \cite{hariharan2011semantic} and Microsoft COCO dataset \cite{lin2014microsoft} together. We use all images in training and validation sets with at least one object from the Pascal VOC 2012 categories except that from the Pascal VOC 2012 validation set. Objects of other categories were treated as background. We use the PASCAL VOC 2012 validation set to evaluate the improvements gained by the structural changes of networks.\\

Follow the work of \cite{yu2015multi}, we first extend the context networks. In all of our experiments, we begin from FCN-32s, take the combination of the output feature maps from conv4\_3, conv5\_3 and fc7 as input feature maps of the remaining module. We keep the parameters of layers in FCN-32s fixed. The number of channels of feature maps from fc7, conv5\_3, conv4\_3 are 4096, 512, 256. Three added layers reduce the channels to 21 before they are confused. If the parameters of these three layers parameters are copied from FCN-8s, and keep unchanged, the model achieves 76.5\% (meanIU) in Pascal VOC 2012 validation set. If these three layers are trained, we got 1 percent improvement. If concatenating the feature maps of these three layers, and the number of channels is up to 63, we got another 0.5 percent improvement.\\

Further experiments show that, with more number of channels of the input feature maps of MCN, the network tends to learn better, for it can capture more information. To alleviating over-fitting, we set the dimension to 256 in following experiments. The numbers of filters of dilated convolutional layers in context network are set to 256, 256, 512, 512, 1024 and 1024 from bottom to top. Here, we did not double the number of filters layer by layer as \cite{yu2015multi} to avoid too many parameters. In our experiments, this model, referred as FCN\_256CT, provide 0.5 percent improvement, achieves 78.4\%. We use this model as the baseline in the following experiments.\\

Then, we consider four models. In the first model, we add a skip connection between the input and the output of the context network module of FCN\_256CT, referred as FCN\_256CT\_long\_skip (Fig.2 (a)). In the second model, we combine short connections and dilated convolutions in FCN\_256CT, referred as FCN\_256CT\_short\_skip (Fig.2 (b)). In the third model, we use mixed context network to replace the context network, referred as FCN\_MCN. In the fourth model, a long skip connection between the input and the output of mixed context network is added, referred as FCN\_MCN\_long\_skip. The result is shown in Table \ref{Table_1}.\\

The mean IOU of FCN\_MCN and FCN\_MCN\_long\_skip are: 80.6\% and 80.2\%, shows that the feature maps from FCN provide no more useful information, combined with feature map from MCN. However, for the FCN\_256CT, the additional skip connection brings 0.9\% improvement. In PASCAL VOC 2012 test set, we achieve 0.814 using one FCN\_MCN model with multi-scale input images.

\begin{table}[!t]
\renewcommand{\arraystretch}{1.3}
\caption{Results of Validation Set of SceneParse150}
\label{Table_2}
\centering
\begin{tabular}{c|c|c}
\hline
models & MeanIU & PixelAcc\\
\hline
DilatedNet & 0.3231 & 73.55\%\\
\hline
Cascade-DilatedNet & 0.3231 & 74.52\%\\
\hline
FCN\_MCN(ours) & 0.3832 & 78.04\%\\
\hline
FCN\_MCN\_REFINE(ours) & 0.3880 & 78.05\%\\
\hline
FCN\_MCN\_REFINE\_MPN(ours) & 0.4001 & 78.40\%\\
\hline
FCN\_MCN\_REFINE\_MPN + multiscale(ours) & 0.4158 & 80.01\%\\
\hline
\end{tabular}
\end{table}

\subsection{MIT SceneParsing150}
The dataset of MIT SceneParsing150 \cite{zhou2016semantic} contains 20k training images, 2000 validation images and 3352 testing images. There are totally 150 categories, contains 35 stuff classes and 115 discrete objects.\\

We augment the training images by randomly flipping and randomly scaling (from 0.5 to 1.5), and patches with size of ${448\times448}$ by randomly cutting are fed to the networks.\\

The FCN-32s is first fine-tuned from VGG-16 \cite{simonyan2014very}, then the MCN module is added to the end of the network as described at Fig.\ref{Fig.1}. When training FCN\_MCN, all parameters of layers in FCN-32s are fixed except the fc7 layer. All parameters of new layers are trained and the learning rate of the additional layers multiplies 10. For the FCN\_MCN\_REFINE\_MPN, we set the iteration number of MPN to 3. The results of validation set are shown in Table \ref{Table_2}, and we achieve 0.53355 in the test set.\\

\section{Conclusion}
In this paper, we introduce a module called mixed convolutional network, which can be added to any FCN models. The experiments show that it can provide better features to predict stuff, big objects and small objects all at once. Furthermore, a memory-efficient module called message passing network is proposed to enforce spatial consistency across labels.

\bibliographystyle{IEEEtran}
\bibliography{IEEEabrv,../bib/paper}

\end{document}